\documentclass{article} 
\usepackage{algorithm}
\usepackage{algorithmic}
\usepackage{amsmath,amssymb,amsfonts,verbatim}
\usepackage{amsthm}
\usepackage{hyperref}
\usepackage{url}
\usepackage[square,numbers,sort]{natbib}
\usepackage{graphicx}
\usepackage{subfigure}
\usepackage{authblk}

\newtheorem{theorem}{Theorem}
\newtheorem{lemma}{Lemma}
\newtheorem{assumption}{Assumption}

\newtheorem{definition}{Definition}

\def\X{\mathcal{X}}

\title{A Parallel algorithm for $\X$-Armed bandits}
\author[1]{Cheng Chen\thanks{jack\_chen1990@sjtu.edu.cn}}
\author[1]{Shuang Liu\thanks{liushuang93006@gmail.com}}
\author[1]{Zhihua Zhang\thanks{zhang-zh@cs.sjtu.edu.cn}}
\author[2]{Wu-Jun Li\thanks{liwujun@nju.edu.cn}}
\affil[1]{Shanghai Jiao Tong University}
\affil[2]{Nanjing University}

\begin{document}

\maketitle

\begin{abstract}
The target of $\X$-armed bandit problem is to find the global maximum of an unknown stochastic function $f$, given a finite budget of $n$ evaluations. Recently, $\X$-armed bandits have been widely used in many situations. Many of these applications need to deal with large-scale data sets. To deal with these large-scale data sets, we study a distributed setting of $\X$-armed bandits, where $m$ players collaborate to find the maximum of the unknown function. We develop a novel anytime distributed $\X$-armed bandit algorithm. Compared with  prior work on $\X$-armed bandits, our algorithm uses a quite different searching strategy so as to fit distributed learning scenarios. Our theoretical analysis shows that our distributed algorithm is $m$ times faster than the classical single-player algorithm. Moreover, the number of communication rounds of our algorithm is only logarithmic in $mn$. The numerical results show that our method can make effective use of every players to minimize the loss. Thus, our distributed approach is attractive and useful.
\end{abstract}

\section{Introduction}

We consider the problem of approximately finding the maximum of an unknown stochastic function $f:\X\rightarrow\mathbb{R}$. Assume that every evaluation of the function is expensive. Thus, we are only given a finite budget of evaluations. Furthermore, each evaluation of the function is perturbed by a noise. More precisely, an evaluation of $f$ at $x_t$ returns a noisy estimation $r_t$ such that
\[ \mathbb{E}[r_t|x_t]=f(x_t).
\]
This problem is called $\X$\textit{-armed bandit} because each evaluation of the function can be considered as pulling an arm in a measurable space $\X$.

Recently, 
the $\X$-armed bandit problem has become increasingly popular \cite{auer2007improved,kleinberg2008multi,bubeck2009online,valko2013stochastic,bull2013adaptive,djolonga2013high,azar2014online,munos2013bandits}. Moreover, the $\X$-armed bandit model has been used in many applications such as ranking documents of search engines \cite{slivkins2013ranked} and  planning in Markov Decision Processes \cite{mansley2011sample}. In the big data era, data sizes of many of these applications are growing to an unprecedentedly large scale. On the one hand, the big data brings us a lot of insightful information. On the other hand, it would also bring computational challenges. Specifically,
it is hard to handle these large-scale applications in a single machine. 

Alternatively,
a distributed approach  to the $\X$-armed bandit problem on large-scale data sets is a desirable choice~\cite{agarwal2011distributed,dekel2012optimal,hillel2013distributed}.
In a  distributed setting of $\X$-armed bandits, particularly,
there are $m$ players corresponding to $m$ independent machines,
which pull arms in parallel.
All players share the arm space $\X$. At every time step, each player pulls an arm from the arm space and obtains the corresponding reward. In order to share information, players may communicate with each other. 
The \textit{communication rounds} are set between two time steps. 
Moreover, in a communication round each player is allowed to broadcast a message to all other players.
In other words, each player can obtain information from all other players in a communication round.

Since data flowing through the network might incur latency or delay, players should avoid communicating too frequently; that is, one should limit the number of communication rounds. For an extreme example, if there is a communication round after each time step, one can simply present an $m$-player algorithm that achieves a speed-up of factor $m$, when compared to the serial setting. However, one cannot afford the cost of communications.
This paper aims to find an algorithm with acceptable cost of communications and maximum factor of speed-up.

Due to the cost of communications, we can find that traditional methods for $\X$-armed bandits such as StoSOO algorithm \cite{valko2013stochastic} and HOO algorithm \cite{bubeck2011x} can hardly be put into a distributed framework. These methods all operates in many traversals of a covering tree from the root down to leaves. If two players traverse the covering tree without communication, they cannot know whether they are visiting the same node. Thus, the whole process may be inefficient and uncoordinated unless large number of communication rounds are set in the algorithm.

Motivated by this, we propose a novel distributed algorithm for the $\X$-armed bandit problem with multiple communication rounds. Our algorithm is an anytime algorithm; that is, our algorithm does not require the knowledge of the number of steps $n$. Moreover, our analysis shows that our algorithm only needs $O(\log(mn))$ communication rounds and that the loss of our algorithm is $O((mn)^{-1/(d+2)}(\log n)^{1/(d+2)})$, where $d$ is the near-optimality dimension (see Section \ref{sec:3}). It is worth pointing out  that  in  previous work on $\X$-armed bandits under the serial setting~\cite{bubeck2009online,kleinberg2008multi,valko2013stochastic,azar2014online}, the loss of the algorithms  is $O((n)^{-1/(d+2)}(\log n)^{1/(d+2)})$. This implies that our method attains a speed-up of a factor more than $m$. In other word, in contrast to  prior work under the serial setting, our algorithm only needs $1/m$ times of number of evaluations to achieve the same loss.

The remainder of this paper is organized as follows. In Section~\ref{sec:related} we review some related work about $\X$-armed bandits and distributed stochastic optimization problems. In Section \ref{sec:3}, we formally setup the distributed $\X$-armed bandit problem
and introduce some notation
and assumptions which will be used in our work. We present our algorithm in Section \ref{sec:4} and conduct theoretical analysis in Section \ref{sec:5}. After that, we provide experiments results of our algorithm in Section ~\ref{sec:6}. Finally, we conclude our work in Section~\ref{sec:7}.

\section{Related Work}
\label{sec:related}

The $\X$-armed bandit problem has been recently intensively studied by many researchers. \citet{kleinberg2008multi} proposed the Zooming algorithm for solving $\X$-armed bandits in which payoff function satisfies a Lipschitz condition. Then, \citet{bubeck2009online} presented a tree-based optimization method called HOO. In contrast with the Zooming algorithm, HOO is an anytime algorithm and able to deal with high order smoothness of the objective function. HOO constructs a covering tree to explore the arm space. This structure is also used in many other algorithms in $\X$-armed bandits, such as the StoSOO algorithm \cite{valko2013stochastic} and the HCT algorithm \cite{azar2014online}. However, all of these methods are not applicable in the distributed framework.

On the other hand, distributed stochastic optimization \cite{agarwal2011distributed,dekel2012optimal,hillel2013distributed} and distributed PAC models \cite{daume2012protocols,daume2012efficient,balcan2012distributed} have also been intensively studied. Among them, \citet{hillel2013distributed} proposed the distributed algorithm for multi-armed bandits. They studied two kinds of distributed algorithms for multi-armed bandits and discussed the tradeoff between the learning performance and the cost of communication. Inspired by their idea, we propose the distributed algorithm for the $\X$-armed bandit problem. In contrast to \citet{hillel2013distributed}'s study in which the arm space is finite and discrete,
our method is the first to study distributed $\X$-armed bandit problem in which the arm space could be any measurable space.

\section{Problem and Assumptions}
\label{sec:3}

We assume that there are $m$ players in the distributed $\X$-armed bandit problem. These players are all given the arm space $\X$. At every time step $t= 1,2,\dots,n$, each player evaluates a point $x_t\in\X$ of his own choice and obtains an independent reward $r_t\in[0,1]$ such that
$$\mathbb{E}[r_t|x_t]=f(x_t).$$
During this process, there are several communication rounds. These communication rounds may take place between any two adjacent time steps. In a communication round, each player is allowed to broadcast a message to all other players. After all players have performed $n$ evaluations, the algorithm outputs a point $x(n)$. We assume that the unknown function $f$ has at least one global maximum $f^*=\sup_{x\in\X}f(x)$ and denote the corresponding maximizer by $x^*=\arg\max_{x\in\X}f(x)$. Then, the performance of a distributed algorithm can be evaluated by the \textit{loss}:
\begin{equation} \label{eq1}
R_n=f^*-f(x(n)).
\end{equation}

\subsection{The Covering Tree}

Similar to recent $\X$-armed bandits methods \cite{bubeck2011x,valko2013stochastic,azar2014online}, our distributed algorithm in this paper also tries to find the maximum by building a binary covering tree $\mathcal{T}$, in which each node covers a subset of $\X$ (called cell). The nodes in the tree are organized according to their depths $h\geq0$ and the depth of the root node is $h=0$. Moreover, nodes at depth $h$ are indexed by $1\leq i\leq 2^h$. We let $(h,i)$ denote the $i$-th node at depth $h$ and index $i$, and $\X_{h,i}\subseteq\X$ denote the corresponding cell. We use $\mathcal{L}_{h,i}$ for the leaves of node $(h,i)$. In addition, each node $(h,i)$ is assigned a representative point $x_{h,i}\in\X_{h,i}$. More on this subject can be found in \cite{bubeck2011x,munos2011optimistic}.

\subsection{Assumptions}

We now make four assumptions which have been also used by \citet{bubeck2011x}. The first two assumptions are about a semi-metric function $\ell$ and the local smoothness of $f$ w.r.t.\ $\ell$, while the last two assumptions are about the structure of the hierarchical partition w.r.t.\ $\ell$.
\begin{assumption}[Semi-metric] \label{assumption:1}
We assume that the space $\X$ is equipped with a function $\ell:\mathcal{X}\times\mathcal{X}\rightarrow\mathbb{R}_+$ such that for all $x, y\in\X$, we have $\ell(x,y)=\ell(y,x)$ and $\ell(x,y)=0$ if and only if $x=y$.
\end{assumption}
\begin{assumption}[Local smoothness of $f$] \label{assumption:2}
For all $x\in\mathcal{X}$, we have $f^*-f(x)\leq\ell(x, x^*)$.
\end{assumption}
\begin{assumption}[Bounded diameters] \label{assumption:3}
We assume that there exists constants $\nu_1>0$ and $0<\rho<1$, such that for any cell $\mathcal{X}_{h,i}$ of depth $h$, we have $\sup_{x\in\mathcal{X}_{h,i}}\ell(x_{h,i},x)\leq \nu_1\rho^h$.
\end{assumption}
\begin{assumption}[Well-shaped cells] \label{assumption:4}
There exists a $\nu_2>0$ such that for any depth $h\geq0$, any cell $\mathcal{X}_{h,i}$ contains a $\ell$-ball of radius $\nu_2 \rho^h$ centered in $x_{h,i}$.
\end{assumption}

\section{Algorithm} \label{sec:4}

We now present the distributed algorithm for  $\X$-armed bandits. Our algorithm is given in Algorithm \ref{alg:1}. We leverage a covering tree $\mathcal{T}$ which is initialized with only a root node $(0,1)$. Our algorithm traverses the tree in level-order. To decide which nodes should be evaluated, we build a confidence set $S_h$ for every depth $h$. $S_h$ contains nodes at depth $h$ that are likely to contain the optimal point. For each depth $h$, the algorithm evaluates nodes in $S_h$ and expands some of them. Then, we use the successors of these expanded nodes at depth $h$ to make up the confidence set $S_{h+1}$ for the next level.

\begin{algorithm}[tb]
   \caption{Distributed algorithm for $\X$-armed bandits.}
   \label{alg:1}
\begin{algorithmic}
   \STATE {\bfseries Input:} number of players $m$, and $\delta>0$.
   \STATE {\bfseries Initialization:}
   \STATE $\ \ \ \ S_0\leftarrow\{(0,1)\}$ \{root node\}
   \STATE $\ \ \ \ h\leftarrow 0$ \{current depth\}
   \STATE $\ \ \ \ t\leftarrow 0$ \{time step\}
   \STATE $\ \ \ \ T_0\leftarrow\frac{\log(\pi^2/3\delta)}{2\nu_1m}$
   \LOOP
   \FOR {player $j=1$ to $m$}
   \FOR {$(h,i)\in S_h$}
   \FOR {$l=1$ to $T_h$}
   \STATE evaluate $x_{h,i}$ and observe the reward $r^{j,l}(x_{h,i})$
   \IF {$j=1$}
   \STATE $t\leftarrow t+1$
   \ENDIF
   \ENDFOR
   \STATE $\hat{\mu}^j_{h,i}\leftarrow\frac{1}{T_h}\sum\limits_{l=1}^{T_h}r^{j,l}(x_{h,i})$
   \ENDFOR
   \STATE communicate mean rewards $\hat{\mu}^j_{h,i}$ for all $(h,i)\in S_h$
   \ENDFOR
   \STATE let $\hat{\mu}_{h,i}=\frac{1}{m}\sum^m_{j=1}\hat{\mu}^j_{h,i}$ for all $(h,i)\in S_h$
   \STATE $\hat{\mu}^*_h\leftarrow \max_{(h,i)\in S_h}\hat{\mu}_{h,i}$
   \STATE $S_{h+1}=\emptyset$
   \FOR {$(h,i)\in S_h$}
   \IF {$\hat{\mu}_{h,i} \geq \hat{\mu}^*_h-3\nu_1\rho^h$}
   \STATE $S_{h+1}\leftarrow S_{h+1}\cup \mathcal{L}_{h,i}$
   \ENDIF
   \ENDFOR
   \STATE $h\leftarrow h+1$
   \STATE $T_h\leftarrow\lceil\frac{\log(\pi^2(h+1)^2|S_h|/3\delta)}{2(\nu_1\rho^h)^2m}\rceil$
   \ENDLOOP
   \STATE let $h_{max}$ be the depth of the deepest expanded node
   \STATE {\bfseries Output:} $x(n)=\arg\max\limits_{x_{h_{max},i}}\hat{\mu}_{h_{max},i}$
\end{algorithmic}
\end{algorithm}

Since  evaluations are all disturbed by noise, it is essential to evaluate every selected node $(h,i)$ for sufficient times to achieve a confidence estimation of $f(x_{h,i})$. In our algorithm, all nodes at depth $h$ are required to be evaluated for the same number of times $mT_h$. Since the evaluation order of nodes in the same level does not affect the result of the algorithm, evaluations at the same depth can be performed in parallel. In our algorithm, for every $(h,i)\in S_h$, each player is required to evaluate the corresponding representative point $x_{h,i}$ for $T_h$ times. The average reward of $x_{h,i}$ that evaluated by player $j$ can be computed as:
$$\hat{\mu}^j_{h,i}=\frac{1}{T_h}\sum\limits_{l=1}^{T_h}r^{j,l}(x_{h,i}),$$
where $r^{j,l}(x_{h,i})$ denotes the $l$-th reward observed by player $j$ after pulling $x_{h,i}$. As showed in the next section, the following choice of $T_h$ is enough for our algorithm:
$$T_h=\lceil\frac{\log(\pi^2(h+1)^2|S_h|/3\delta)}{2(\nu_1\rho^h)^2m}\rceil,$$
where $\delta$ is the confidence parameter. After the players have finished sampling all nodes at depth $h$, they communicate with each other to attain others' mean rewards of nodes in $S_h$. Then, all players can integrate these rewards and obtain accurate confident estimates of nodes that belong to $S_h$. For node $(h,i)$, the corresponding estimate $\hat{\mu}_{h,i}$ is computed as:
$$\hat{\mu}_{h,i}=\frac{1}{m}\sum^m_{j=1}\hat{\mu}^j_{h,i}.$$

Another key ingredient of our algorithm is how to construct the confidence set $S_h$. At one extreme, if we expand all nodes in the tree, we can ensure that nodes containing the maximum point at each depth will not be missed by any players. However, such a strategy is not able to visit nodes at deep levels in the tree because it wastes a large number of evaluations on suboptimal nodes. At the other extreme, for each depth $h$, if we only expand the leaves with the highest estimate, we can visit deeper nodes. However, we have high probability to miss the node  containing the optimum $x^*$ because we cannot guarantee that the representative point $x_{h,i}$ is the optimal point in the cell $\X_{h,i}$.
This yields an exploration-exploitation tradeoff; i.e., the exploration is to sample more nodes at the same depth and the exploitation is to sample deeper nodes in the tree.

Our strategy is described as follows. We denote the maximum estimate at depth $h$ as $\hat{\mu}^*_h$, namely,
\[ \hat{\mu}^*_h= \max_{(h,i)\in S_h}\hat{\mu}_{h,i}.
\]
In our algorithm, we only expand nodes whose estimates satisfy the following condition:
$$\hat{\mu}_{h,i} \geq \hat{\mu}^*_h-3\nu_1\rho^h.$$
This condition eliminates the suboptimal leaves at each depth $h$. Moreover, with this condition, we can give an accurate definition of the confidence set $S_h$ as follows:
\begin{definition}
Let the confidence set at depth $h$ be the set of all nodes that need to be evaluated by players:
\begin{equation*}
\begin{split}
S_h\stackrel{def}{=}\{\text{nodes } (h,i)\text{ whose parent }(h-1,k) \text{ satisfies }\hat{\mu}_{h-1,k} \geq \hat{\mu}^*_{h-1}-3\nu_1\rho^{h-1}\}.
\end{split}
\end{equation*}
\end{definition}
According to the analysis in Section~\ref{sec:5}, we can guarantee that with high probability nodes at depth $h$ containing the optimum $x^*$ belong to $S_h$. In other words, our algorithm is optimistic.

Our algorithm is terminated once $n$ evaluations for each player are exhausted, namely, the loop in the algorithm finishes when the time step $t$ equals $n$. At the end of the algorithm, we  define $h_{max}$ as the depth of the deepest node  expanded and return the representative point $x_{h_{max},i}$ of the node at the depth $h_{max}$ which has the maximal estimate.

\section{Analysis} \label{sec:5}

In this section we provide the theoretical analysis of our algorithm and put all proofs of lemmas and theorems in the appendix. Firstly, we analyze the upper bound of the loss defined in (\ref{eq1}). Then, we analyze the upper bound of the number of communication rounds in our algorithm. 

\subsection{Upper Bound of the Loss}
In order to bound the loss of our algorithm, we use the similar definition of near-optimality dimension in \cite{bubeck2011x}. For any $\epsilon>0$, we denote the $\epsilon$-optimal set as
\[
H_{\epsilon}\stackrel{def}{=}\{x\in\X: f(x)\geq f^*-\epsilon\}.
\]
\begin{definition} \label{def}
The $\eta$-near-optimality dimension is the smallest $d\geq 0$ such that there exists a $C>0$ such that for any $\varepsilon>0$, the maximum number of disjoint $\ell$-balls of radius $\eta\varepsilon$ and center in $H_{\varepsilon}$ is less than $C\varepsilon^{-d}$.
\end{definition}

In our algorithm, the accuracy of the estimate $\hat{\mu}_{h,i}$ of $x_{h,i}$ is essential to the final result because it directly affects which cells to sample or expand. So, we first give the probability of the event that the average estimate $\hat{\mu}_{h,i}$ of expanded node $(h,i)$ is  very close to the true value $f(x_{h,i})$. The following lemma defines such an event and show that this event happens with high probability.

\begin{lemma}\label{lemma:1}
We define the event as follows:
\[
\xi=\bigg\{\forall h\geq 0,\forall (h,i)\in S_h,|\hat{\mu}_{h,i}-f(x_{h,i})|\leq \sqrt{\frac{\log(\pi^2(h+1)^2|S_h|/3\delta)}{2mT_h}}\bigg\}.
\]
Then, the event $\xi$ holds with probability at least $1-\delta$.
\end{lemma}

Lemma \ref{lemma:1} shows that when the leaf node $(h,i)$ is evaluated for $mT_h$ times, the mean estimate $\hat{\mu}_{h,i}$ is very close to the true value $f(x_{h,i})$ with high probability. Note that in  previous work on $\X$-armed bandits such as HOO \cite{bubeck2009online}, StoSOO \cite{valko2013stochastic} and HCT \cite{azar2014online}, they need to guarantee the accuracy of mean estimates at any time $t$. However, in the event $\xi$, we only consider the situation where all $mT_h$ evaluations have been performed for the node $(h,i)$. The reason is that the estimate $\hat{\mu}_{h,i}$ can be only obtained after all players communicate their evaluations of $x_{h,i}$. Considering the mean reward at any time $t$ is meaningless in our algorithm.

For the convenience of the following analysis, we let $$\varepsilon=\sqrt{\frac{\log(\pi^2(h+1)^2|S_h|/3\delta)}{2mT_h}}.$$
Then, Lemma \ref{lemma:1} shows that with probability $1-\delta$, for all $h\geq0$ and for all $(h,i)\in S_h$,  we have:
$$|\hat{\mu}_{h,i}-f(x_{h,i})|\leq\varepsilon.$$
In addition, according to the value of $T_h$ used in our algorithm, we can obtain that $\varepsilon\leq\nu_1\rho^h$.

The next lemma shows what kind of nodes will be involved in the confidence set $S_h$.
\begin{lemma} \label{lemma:2}
In the event $\xi$, for any node $(h,i)$ that satisfies
\begin{equation*} 
f(x_{h,i})\geq f^*-\nu_1\rho^h,
\end{equation*}
its leaves belong to the set $S_{h+1}$.
\end{lemma}
Let $(h,i_*)$  denote the node which is at depth $h$ and satisfies $x^*\in\X_{h,i_*}$. Then, in Lemma~\ref{lemma:3}, we show that with high probability these nodes containing the maximum $x^*$ are selected by our algorithm.
\begin{lemma} \label{lemma:3}
In the event $\xi$, for any depth $h$, the node $(h,i_*)$ will be put into $S_h$ by our algorithm; i.e., these nodes will be evaluated by players.
\end{lemma}

Lemma \ref{lemma:3} is very important for us because it promises that our algorithm will not miss the optimal leaves when we continuously go deep in the covering tree. This means that our algorithm is optimistic. Therefore, our main theorems are all based on this lemma.

The following lemma shows the characteristics of the nodes which are expanded by our algorithm.
\begin{lemma} \label{lemma:4}
In the event $\xi$, for all nodes $(h,i)$ at depth $h$ that are expanded, we have $$f(x_{h,i})+6\nu_1\rho^h\geq f^*.$$
\end{lemma}

Lemma \ref{lemma:4} shows a restriction of expanded nodes. It says that we only need to expand a very small part of nodes in the tree. This lemma guarantees the efficiency of our algorithm. In order to know exactly how many elements need to be evaluated, we bound the size of $S_h$ at depth $h$ in the following lemma.
\begin{lemma} \label{lemma:5}
Let $d$ be the $\frac{\nu_2}{6\nu_1}$-near-optimality dimension, and $C$ be the corresponding constant. Then, in the event $\xi$, for all $h\geq 1$, the cardinality of the confidence set $S_h$ at depth $h$ is bounded as:
$$|S_h|\leq 2C(6\nu_1\rho^{h-1})^{-d}.$$
\end{lemma}

Note that Lemma \ref{lemma:5} does not bound the number of nodes in $S_0$, but we can simply get that $|S_0|=1$ according to our algorithm. Therefore, the cardinality of the confidence set at each depth has an upper bound. These upper bounds ensure that our algorithm is able to visit deep nodes in the tree rather than waste evaluations on suboptimal nodes. Therefore, using these upper bounds, the following lemma gives a lower bound of the depth of the deepest nodes.

\begin{lemma} \label{thm:1}
With probability $1-\delta$, the depth of the deepest expanded nodes $h_{max}$ is bounded as:
$$h_{max}\geq\log_{\rho}\left(c_1\left[\frac{\log(\pi^2n^3/3\delta)}{mn}\right]^{\frac{1}{d+2}}\right),$$
where $c_1=\frac{1}{\rho\nu_1}(\frac{C6^{-d}}{1-\rho^{d+2}})^{\frac{1}{d+2}}$.
\end{lemma}

Lemma \ref{thm:1} provides us a lower bound of $h_{max}$. Since we will not miss the optimal nodes at each level, the deeper nodes in the tree we reach, the more accurate results we can achieve. Therefore, we can take advantage of the bound of $h_{max}$ to bound the loss $R_n$ of our algorithm. We now present our main theorem:

\begin{theorem} \label{thm:2}
With probability $1-\delta$, the loss of our algorithm is bounded as
$$R_n\leq O\left(\left(\frac{\log(n/\delta)}{mn}\right)^{\frac{1}{d+2}}\right).$$
\end{theorem}
Theorem \ref{thm:2} provides an upper bound of the loss $R_n$. For the choice $\delta=\frac{1}{n}$, we have
$$R_n\leq O\left(\left(\frac{\log(n)}{mn}\right)^{\frac{1}{d+2}}\right).$$

Since the loss of $\X$-armed bandits under the serial setting is $O((n)^{-1/(d+2)}(\log n)^{1/(d+2)})$, we can note that our distributed algorithm speeds up more than $m$ times. More precisely, in contrast to  prior work under the serial setting, our algorithm only needs $1/m$ times of number of evaluations to achieve the same loss. If we set $m=1$, we can find that the loss of our algorithm is the same as the loss under the serial setting.

\subsection{Upper Bound of the Communication Cost}

In this part, we consider the cost of communications in two aspects. The number of communication rounds is the major factor that influences the communication cost because sending and receiving messages cost much time. Another factor about the communication cost is the total amount of communication of a single player. Here we will show that both the number of communication rounds and the amount of communication  have upper bounds. Thus, the communication cost is also bounded.

We first discuss the number of communication rounds. The following theorem gives an upper bound of the number of communication rounds.
\begin{theorem} \label{thm:3}
The number of communication rounds $q$ satisfies
$$q\leq O(\log(mn)). $$
\end{theorem}

Theorem \ref{thm:3} shows that our algorithm only need to communicate for only logarithmic to $mn$ times. This result is tolerable because the number of communication rounds is quite small when it is compared to the number of evaluations $n$.

Finally, we report the bound on the total amount of the communication of a single player.

\begin{theorem} \label{thm:4}
Assume that each player broadcasts $M$ values through the algorithm. Then, with probability $1-\delta$, $M$ is bounded as:
$$M\leq O\left([\frac{mn}{\log n}]^{\frac{d}{d+2}}\right).$$
\end{theorem}

\section{Experiments}
\label{sec:6}

\begin{figure} [ht]
\centering
\subfigure[]{
\label{sinf}
\includegraphics[width = 0.45\textwidth]{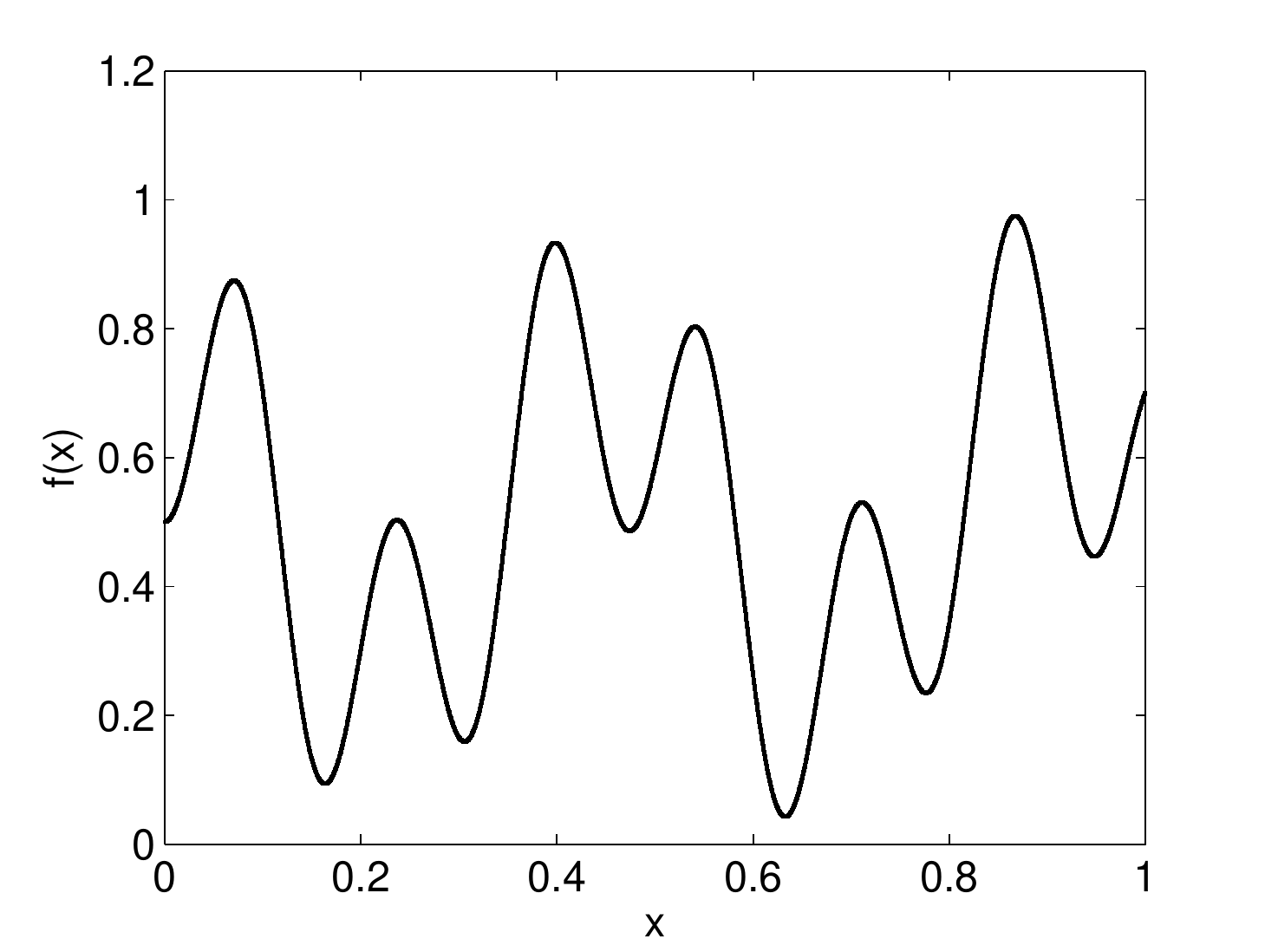}
}
\hspace{1mm}
\subfigure[]{
\label{garf}
\includegraphics[width = 0.45\textwidth]{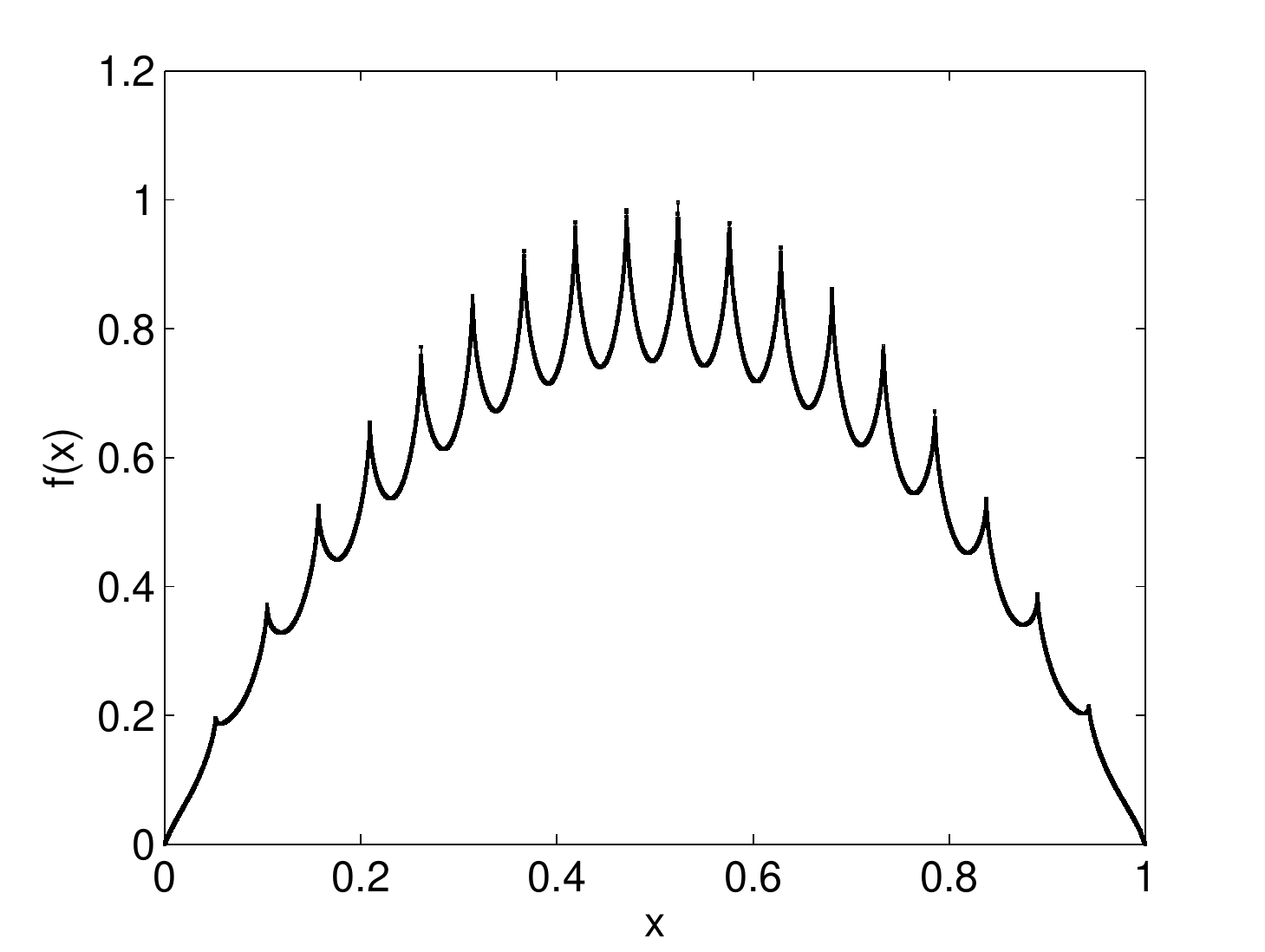}
}
\caption{Test Functions. (a): $f(x)=\frac{1}{2}(\sin(13x)\sin(27x)/2+1)$. (b): The garland function.} \label{garf}
\end{figure}

In this section we provide the numerical evaluation of the performance of our algorithm. In our experiments, we focus on searching the global optima of two functions. The first function is $f(x)=\frac{1}{2}(\sin(13x)\sin(27x)/2+1)$, which is showed in Figure \ref{sinf}. The near-optimality dimension of this function is $d=0$ if we use $\ell_2$-metric. The second function is the garland function $f(x)=x(1-x)(4-\sqrt{|\sin(60x)|})$, which is showed in Figure \ref{garf}. It's near-optimality dimension can also be $d=0$. However, optimization of the garland function is more difficult because it contains much more local optima.

\begin{figure} [ht]
\centering
    \begin{tabular}{cc}
        \includegraphics[width = 0.45\textwidth]{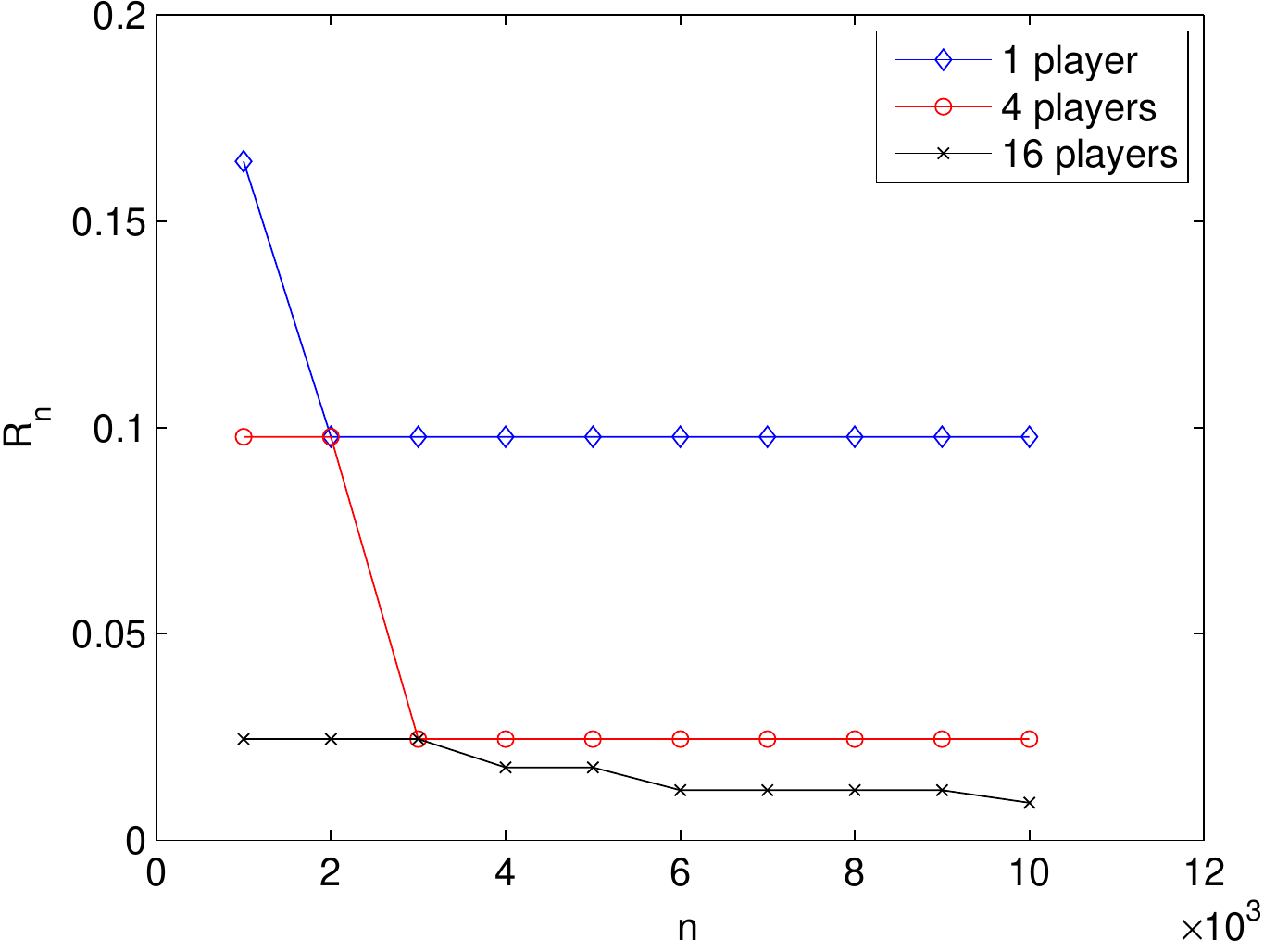} &
        \includegraphics[width = 0.45\textwidth]{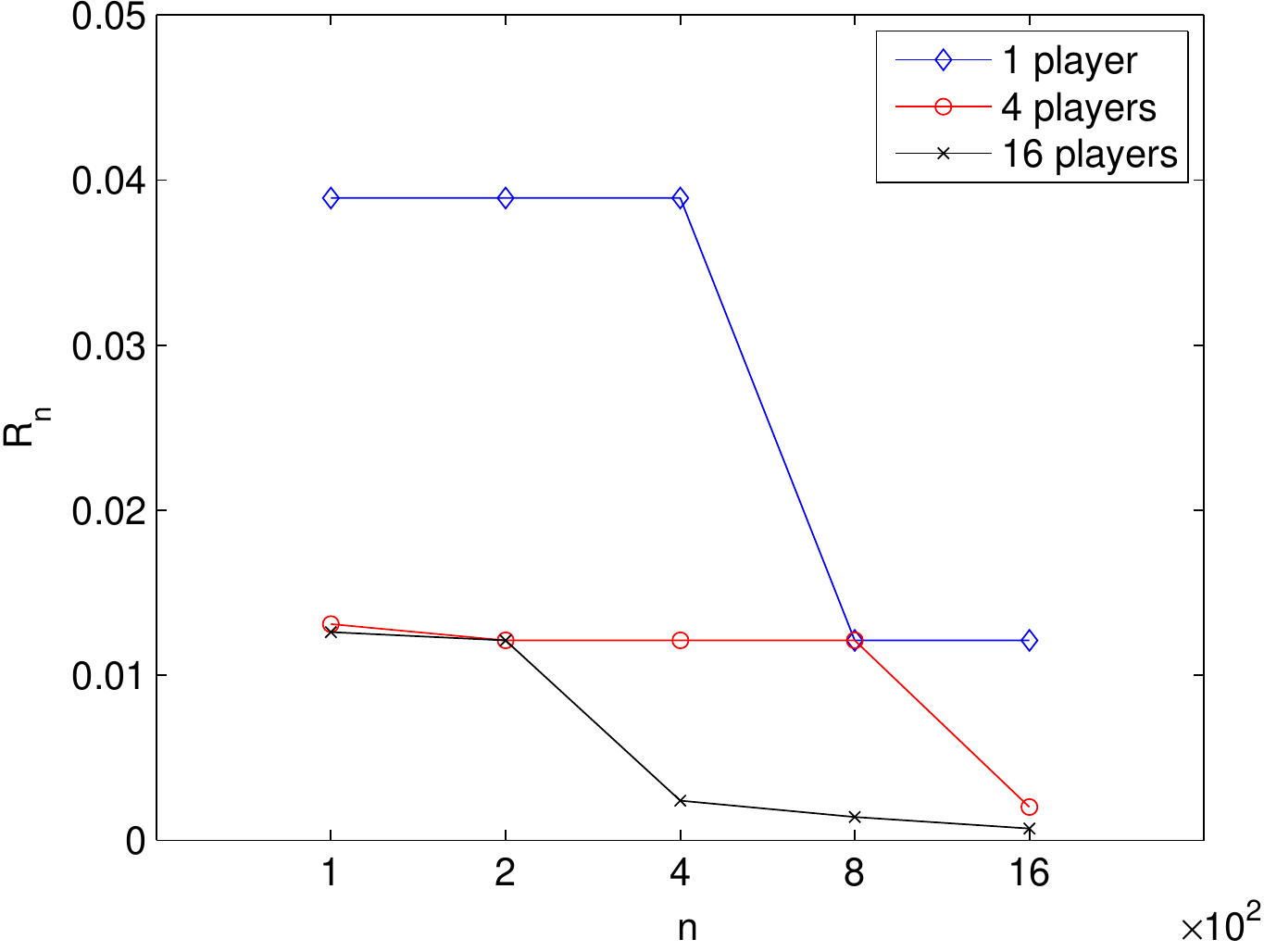}
    \end{tabular}
    \caption{Loss of the distributed $\X$-armed bandits. The left figure is related to the function $f(x)=\frac{1}{2}(\sin(13x)\sin(27x)/2+1)$ and the right figure is related to the garland function.} \label{loss}
\end{figure}

In $\X$-armed bandits settings, all evaluations are noised, i.e., pulling an arm $x$ produces a reward $f(x)+\epsilon$. Therefore, we add  a truncated zero mean Gaussian noise $\mathcal{N}(0,1)$ for each evaluation in our experiments. We truncate the Gaussian distribution because the reward should be bounded in $[0,1]$. We run our algorithm with three settings: 1 player, 4 players and 16 players. Since our algorithm is the first distributed algorithm for $\X$-armed bandits, we compare the loss between different settings. We run $1600$ steps for the first function and run $10000$ steps for the second function. We compare the results of these settings in Figure \ref{loss}.

In the experiment, we do not consider the communication cost because it is hard to compare the cost of one evaluation with the cost of one communication. Thus, the experiment only shows the speed-up factor of our distributed algorithm. On the other hand, the number of communication rounds is rather small so that we can directly ignore the cost of communication in most situations.

As described by the theoretical analysis, our distributed algorithm has an obvious speed-up related to the number of players. We can see that the result of 4 players  outperforms the result of 1 player by a large margin and  the result of 16 players also outperforms the result of 4 players.

\section{Conclusion and Future Work} \label{sec:7}

In this paper we have developed a novel distributed algorithm for $\X$-armed bandits.
In our algorithm, $m$ players collaborate to explore the arm space $\X$ to find the optimum of an unknown function.
We have derived the speed-up factor of our distributed algorithm and presented the bound on the communication cost.

Compared to previous tree-based methods for $\X$-armed bandits~\cite{bubeck2009online,valko2013stochastic,azar2014online}, our main contribution is that we provided a novel search strategy to traverse the partition tree. In particular, we visit nodes in the partition tree in level-order. Our method is particularly suited to a distributed framework because for any depth $h$, we have to sample all nodes that may contain the optimal point at depth $h$ before sampling nodes at depth $h+1$.  The analysis shows that our method achieves a speed-up of a factor more than $m$.

In the future, we plan to develop a distributed algorithm for $\X$-armed bandits that can handle unknown smoothness, i.e., we do not need to know $\nu_1$ and $\rho$ in the algorithm. \citet{valko2013stochastic} provided an algorithm for $\X$-armed bandits with unknown smoothness but their method can not be performed in parallel.

Another interesting problem is to develop a distributed algorithm to minimize the cumulative regret of $\X$-armed bandits. In this setting, each player will produce a regret and we aim to minimize the total amount of all these regrets. The cumulative regret setting faces a great challenge because all players need to consider the exploration-exploitation tradeoff while exploring the arm space.

If we use $R_{loss}$ to represent the loss of a distributed bandit problem and use $R_{regret}$ to represent the cumulative regret of a distributed bandit problem, we can obtain
$$R_{loss}\leq \frac{1}{mn}R_{regret}. $$
Thus, the cumulative regret setting is more general and useful. Although the methods such as HOO \cite{bubeck2009online} and HCT\cite{azar2014online} can deal with the cumulative regret of $\X$-armed bandits, there is no a distributed framework for the cumulative regret setting.
\newpage
\bibliography{nips2015}
\bibliographystyle{unsrtnat}

\newpage
\appendix
\section{Upper Bound of the Loss}
\subsection{Proof of Lemma \ref{lemma:1}}
\begin{proof}
We consider the complementary event of $\xi$:
\[
\xi^c=\bigg\{\exists h\geq 0, \text{ s.t. } \exists (h,i)\in S_h \text{ that satisfies:}\\
|\hat{\mu}_{h,i}-f(x_{h,i})|> \sqrt{\frac{\log(\pi^2(h+1)^2|S_h|/3\delta)}{2mT_h}} \bigg\}.
\]
Then, we upper bound the probability of $\xi^c$ as follows:
\begin{equation*}
\begin{split}
\Pr(\xi^c)&\leq \sum_{h=0}^{h_{max}}\sum_{(h,i)\in S_h}\Pr\left\{|\hat{\mu}_{h,i}-f(x_{h,i})|> \sqrt{\frac{\log(\pi^2(h+1)^2|S_h|/3\delta)}{2mT_h}}\right\}\\
&\leq \sum_{h=0}^{h_{max}}|S_h|\frac{6\delta}{\pi^2(h+1)^2|S_h|}\\
&\leq \frac{6\delta}{\pi^2}\sum_{h=0}^{\infty}\frac{1}{(h+1)^2}\\
&=\delta.
\end{split}
\end{equation*}
The first inequality is an application of a union bound over all expand nodes and the second inequality follows from the Chernoff-Hoeffding inequality. Then, we can achieve that $\Pr(\xi)\geq1-\delta$.
\end{proof}

\subsection{Proof of Lemma \ref{lemma:2}}
\begin{proof}
According to Lemma \ref{lemma:1}, in the event $\xi$, we have
\begin{equation} \label{eq11}
\hat\mu_{h,i}\geq f(x_{h,i})-\varepsilon.
\end{equation}
We denote by $(h^*,i^*)$ the node whose mean estimate is $\hat{\mu}^*_h$. Then, according to \eqref{eq:12}, we can achieve that:
\begin{equation} \label{eq2}
f(x_{h,i})\geq f^*-\nu_1\rho^h\geq f(x_{h^*,i^*})-\nu_1\rho^h.
\end{equation}
Using Lemma \ref{lemma:1} again, we obtain:
\begin{equation} \label{eq3}
f(x_{h^*,i^*})\geq \hat\mu^*_h-\varepsilon.
\end{equation}
Combine (\ref{eq11}), (\ref{eq2}) and (\ref{eq3}), we have that
$$\hat\mu_{h,i}\geq \hat\mu^*_h-\nu_1\rho^h-2\varepsilon\geq \hat\mu^*_h-3\nu_1\rho^h.$$
which completes the proof.
\end{proof}

\subsection{Proof of Lemma \ref{lemma:3}}
\begin{proof}
Since the node $(h,i_*)$ contains the optimum $x^*$, its parent node must also contains $x^*$. We assume that its parent node is $(h-1,i'_*)$. Then, using Assumptions 2 and  3, we can obtain that:
$$f(x_{h-1,i'_*})\geq f^*-\nu_1\rho^{h-1}.$$
Then, according to Lemma \ref{lemma:2}, we know that $(h,i*)$ belongs to $S_h$.
\end{proof}

\subsection{Proof of Lemma \ref{lemma:4}}
\begin{proof}
We  use the same notation as in the previous lemma. Then, according to Lemma \ref{lemma:1}, we have:
\begin{equation} \label{eq4}
f(x_{h,i})\geq \hat{\mu}_{h,i}-\varepsilon
\end{equation}
and
\begin{equation} \label{eq5}
\hat{\mu}_{h,i_*}\geq f(x_{h,i_*})-\varepsilon.
\end{equation}
Assuming that node $(h,i)$ is expanded, using Lemma 2 and Lemma 3, we have
$$\hat{\mu}_{h,i}\geq \hat{\mu}^*_h-3\nu_1\rho^h\geq \hat{\mu}_{h,i_*}-3\nu_1\rho^h.$$
The first inequality is based on the condition of expansion and the second inequality follows from the fact that $\hat{\mu}_{h,i_*}$ is the highest estimate at depth $h$.\\
Then, combining this with \eqref{eq4} and \eqref{eq5} and using Assumptions 2 and  3, we have that
\begin{eqnarray*}
f(x_{h,i})& \geq & f(x_{h,i_*})-2\varepsilon-3\nu_1\rho^h\\
&\geq & f^*-2\varepsilon-4\nu_1\rho^h\\
&\geq & f^*-6\nu_1\rho^h.
\end{eqnarray*}
\end{proof}

\subsection{Proof of Lemma \ref{lemma:5}}
\begin{proof}
We first define the expansion set $I_h$ at depth $h$ as follows:
$$I_h=\{\mathrm{node}\  (h,i)\  \mathrm{such\ that} \  f(x_{h,i})+6\nu_1\rho^h\geq f^*\}.$$
Then, from Lemma \ref{lemma:4}, we can find that $I_h$ contains all nodes at depth $h$ that are expanded. Since our partition tree is a binary tree, we have
$$|S_h|\leq 2|I_{h-1}|.$$
Then, we prove this lemma by contradiction. Assume that for some $h$, we have $|S_h|>2C(6\nu_1\rho^{h-1})^{-d}$. This would imply that $|I_{h-1}|$ exceeds $C(6\nu_1\rho^{h-1})^{-d}$. According to Assumption 4, each cell $\X_{h-1,i}$ contains a $\ell$-ball of radius $\nu_2\rho^{h-1}$ centered in $x_{h-1,i}$. Thus we can obtain that there exists more than $C(6\nu_1\rho^{h-1})^{-d}$ disjoint $\ell$-balls of radius $\nu_2\rho^{h-1}$ with center in $H_{6\nu_1\rho^{h-1}}$.Then, we have a contradiction with the definition of $\frac{\nu_2}{6\nu_1}$-near-optimality dimension $d$.
\end{proof}

\subsection{Proof of Lemma \ref{thm:1}}
\begin{proof}
We consider the total number of evaluations $n$. At depth $h$, each player needs to sample every node in $S_h$ for $T_h$ times. Therefore, each player performs $T_h|S_h|$ evaluations at depth $h$. Since the nodes at depth $h_{max}+1$ are not all evaluated for $T_{h+1}$ times, we have
$$n\leq \sum_{h=0}^{h_{max}+1}T_h|S_h|.$$
Then, using Lemma \ref{lemma:5}, we obtain that
$$n\leq T_0+\sum_{h=1}^{h_{max}+1}\frac{\log(\pi^2(h+1)^2|S_h|/3\delta)}{(\nu_1\rho^h)^2m}C(6\nu_1\rho^{h-1})^{-d}.$$
Using the following inequality: $h+1\leq n$ , $|S_h|\leq n$ , and $2C(6\nu_1)^{-d}\geq |S_1|\geq 1$, we can get
\begin{equation*}
\begin{split}
n&\leq T_0+\sum_{h=1}^{h_{max}+1}\frac{\log(\pi^2n^3/3\delta)}{(\nu_1\rho^h)^2m}C(6\nu_1\rho^h)^{-d}\\
&\leq \sum_{h=0}^{h_{max}+1}\frac{\log(\pi^2n^3/3\delta)}{(\nu_1\rho^h)^2m}C(6\nu_1\rho^h)^{-d}\\
&= \sum_{h=0}^{h_{max}+1}\frac{\log(\pi^2n^3/3\delta)}{(\nu_1\rho^h)^{d+2}m}C6^{-d}\\
&=\frac{\log(\pi^2n^3/3\delta)C6^{-d}}{m}\sum_{h=0}^{h_{max}+1}(\nu_1\rho^h)^{-(d+2)}\\
&=\frac{\log(\pi^2n^3/3\delta)C6^{-d}\nu_1^{-(d+2)}}{m}\cdot\frac{\rho^{-(h_{max}+2)(d+2)}-1}{\rho^{-(d+2)}-1}\\
&\leq \frac{\log(\pi^2n^3/3\delta)C6^{-d}\nu_1^{-(d+2)}}{m}\cdot\frac{\rho^{-(h_{max}+1)(d+2)}}{1-\rho^{d+2}}\\
&= \frac{\log(\pi^2n^3/3\delta)C6^{-d}(\nu_1\rho)^{-(d+2)}}{m(1-\rho^{d+2})}\rho^{-(d+2)h_{max}}\\
&=c_1^{d+2}\rho^{-(d+2)h_{max}}\frac{\log(\pi^2n^3/3\delta)}{m}.
\end{split}
\end{equation*}
Since $0<\rho<1$, we can get
$$h_{max}\geq\log_{\rho}\left(c_1\left[\frac{\log(\pi^2n^3/3\delta)}{mn}\right]^{\frac{1}{d+2}}\right).$$
\end{proof}

\subsection{Proof of Theorem \ref{thm:2}}
\begin{proof}
According to the definition of $x(n)$, we know that $x(n)$ is at depth $h_{max}$. Then, using Lemma \ref{lemma:4}, we can obtain
$$f(x(n))+6\nu_1\rho^{h_{max}}\geq f^*,$$
namely,
$$R_n\leq 6\nu_1\rho^{h_{max}}.$$
According to Lemma 6, we can know that with probability $1-\delta$,
$$R_n\leq 6\nu_1\rho^{h_{max}}\leq 6\nu_1c_1\left[\frac{\log(\pi^2n^3/3\delta)}{mn}\right]^{\frac{1}{d+2}}.$$
Thus, with probability $1-\delta$, we can get
$$R_n\leq O\left(\left(\frac{\log(n/\delta)}{mn}\right)^{\frac{1}{d+2}}\right).$$
\end{proof}

\section{Upper Bound of the Communication Cost}
\subsection{Proof of Theorem \ref{thm:3}}
\begin{proof}
We first show that the depth of the deepest expanded nodes is bounded as:
$$\sum_{h=0}^{h_{max}}\frac{1}{(\nu_1\rho^h)^2}\leq mn.$$
Similar to the proof of Lemma \ref{thm:1}, we consider the total number of evaluations n. For any $h\leq h_{max}$, each node in the confidence set $S_h$ is evaluated for $T_h$ times by all players. Therefore, we have
$$n\geq \sum_{h=0}^{h_{max}}T_h|S_h|.$$
Since we have $|S_h|\geq 1$ and $\frac{\log(\pi^2(h+1)^2|S_h|/3\delta)}{2}\geq 1$, we can get
\begin{equation*}
\begin{split}
n&\geq \sum_{h=0}^{h_{max}}T_h\\
&\geq \sum_{h=0}^{h_{max}}\frac{\log(\pi^2(h+1)^2|S_h|/3\delta)}{2(\nu_1\rho^h)^2m}\\
&\geq \sum_{h=0}^{h_{max}}\frac{1}{(\nu_1\rho^h)^2m}.\\
\end{split}
\end{equation*}
Then, we bound the number of communication rounds $q$. Actually, players only communicate with each other when they finish sampling all nodes in a confidence set of a specific depth. So, the number of communication rounds $q$ is equal to the depth of the deepest node, i.e., $q=h_{max}$. Therefore, using the upper bound of $h_{max}$, we can get
\begin{equation*}
\begin{split}
mn&\geq \sum_{h=0}^{q}\frac{1}{\nu_1^2\rho^{2h}}\\
&= \frac{1}{\nu_1^2}\frac{\rho^{-2(q+1)}-1}{\rho^{-2}-1}\\
&\geq \frac{\rho^{-2q}}{\nu_1^2}.
\end{split}
\end{equation*}
Since $0<\rho<1$, we have
$$q\leq O(\log(mn)).$$
\end{proof}

\subsection{Proof of Theorem \ref{thm:4}}
\begin{proof}
At each depth $h$, each player needs to broadcast the mean rewards of all nodes in $S_h$. Therefore, at depth $h$, each player needs to broadcast $|S_h|$ values.
Then, we have
$$M = \sum_{h=0}^{h_{max}}|S_h|.$$
Using Lemma \ref{lemma:5}, we have that
\begin{equation*}
\begin{split}
M&\leq 1+\sum_{h=1}^{h_{max}}CK(6\nu_1\rho^{h-1})^{-d}\\
&\leq 1+\sum_{h=0}^{h_{max}-1} CK(6\nu_1\rho^h)^{-d}\\
&=1+CK(6\nu_1)^{-d}\frac{\rho^{-dh_{max}}-1}{\rho^{-d}-1}.
\end{split}
\end{equation*}
According to Lemma \ref{thm:1}, we have
$$h_{max}\geq\log_{\rho}\left(c_1\left[\frac{\log(\pi^2n^3/3\delta)}{mn}\right]^{\frac{1}{d+2}}\right).$$
Thus, we can obtain
$$M\leq O\left([\frac{mn}{\log n}]^{\frac{d}{d+2}}\right).$$
\end{proof}

\end{document}